\documentclass{article}

\usepackage{microtype}
\usepackage{graphicx}
\usepackage{subcaption}
\usepackage{booktabs} 

\usepackage{hyperref}

\newcommand{\our}[0]{ProtoQuant}

\usepackage[preprint]{icml2026}

\usepackage{amsmath}
\usepackage{amssymb}
\usepackage{mathtools}
\usepackage{amsthm}
\usepackage{pgfplots}
\pgfplotsset{compat=1.18}
\definecolor{myblue}{RGB}{0, 114, 178}
\definecolor{myred}{RGB}{213, 94, 0}
\definecolor{mygray}{RGB}{128, 128, 128}

\usepackage[capitalize,noabbrev]{cleveref}

\theoremstyle{plain}

\theoremstyle{definition}

\theoremstyle{remark}

\usepackage[textsize=tiny]{todonotes}

\icmltitlerunning{ProtoQuant: Quantization of Prototypical Parts For General and Fine-Grained Image Classification}

\begin{document}

\twocolumn[
  \icmltitle{ProtoQuant: Quantization of Prototypical Parts For General and Fine-Grained Image Classification}




  \begin{icmlauthorlist}
    \icmlauthor{Mikołaj Janusz}{1}
    \icmlauthor{Adam Wróbel}{1,2}
    \icmlauthor{Bartosz Zieliński}{1}
    \icmlauthor{Dawid Rymarczyk}{1,3}
  \end{icmlauthorlist}

  \icmlaffiliation{1}{Jagiellonian University, Faculty of Mathematics and Computer Science}
  \icmlaffiliation{2}{Jagiellonian University, Doctoral School of Exact and Natural Sciences}
  \icmlaffiliation{3}{Ardigen SA}

  \icmlcorrespondingauthor{Dawid Rymarczyk}{dawid.rymarczyk@uj.edu.pl}

  \icmlkeywords{prototypical parts, quantization, interpretability}

  \vskip 0.3in
]



\printAffiliationsAndNotice{}  

\begin{abstract}

Prototypical parts-based models offer a "this looks like that" paradigm for intrinsic interpretability, yet they typically struggle with ImageNet-scale generalization and often require computationally expensive backbone finetuning. Furthermore, existing methods frequently suffer from "prototype drift," where learned prototypes lack tangible grounding in the training distribution and change their activation under small perturbations. We present ProtoQuant, a novel architecture that achieves prototype stability and grounded interpretability through latent vector quantization. By constraining prototypes to a discrete learned codebook within the latent space, we ensure they remain faithful representations of the training data without the need to update the backbone. This design allows ProtoQuant to function as an efficient, interpretable head that scales to large-scale datasets. We evaluate ProtoQuant on ImageNet and several fine-grained benchmarks (CUB-200, Cars-196). Our results demonstrate that ProtoQuant achieves competitive classification accuracy while generalizing to ImageNet and comparable interpretability metrics to other prototypical-parts-based methods.

\end{abstract}

\begin{table*}[t]
\centering
\caption{Comparison of prototypical part-based models across key architectural and functional characteristics. ProtoQuant is the only method that follows all of them.}
\label{tab:model_comparison}
\small
\begin{tabular}{lcccc}
\toprule
\textbf{Model} & \textbf{Grounded Proto.} & \textbf{No Backbone Tuning} & \textbf{Backbone Agnostic} & \textbf{Generalizes to ImageNet} \\ \midrule
ProtoPNet~\cite{chen2019looks}      & \checkmark &   &   &   \\
ProtoViT~\cite{ma2024interpretable}       & \checkmark &   &   &   \\
PIPNet~\cite{nauta2023pip}         &   &   &   &   \\
MCPNet~\cite{wang2024mcpnet}         &   &   &   &   \\
PixPNet~\cite{carmichael2024pixel} & \checkmark  &   &   &   \\
ProtoPFormer~\cite{xue2022protopformer} &   &   &   &   \\
ProtoS-ViT~\cite{turbe2024protos}     & \checkmark &  \checkmark &   &  \\
InfoDisent~\cite{struski2024infodisent}     &   & \checkmark & \checkmark & \checkmark \\
EPIC~\cite{borycki2025epic}           &   & \checkmark & \checkmark & \checkmark \\
ProtoQuant (Ours)    & \textbf{\checkmark} & \textbf{\checkmark} & \textbf{\checkmark} & \textbf{\checkmark} \\ 
\bottomrule
\end{tabular}
\vspace{-1em}
\end{table*}

\section{Introduction}

Deep learning (DL) models achieved remarkable success across a diverse computer vision tasks~\cite{khan2022transformers}. However, their deployment in high-stakes domains, such as medical diagnostics and autonomous driving, remains constrained by the inherent opaqueness of traditional DL architectures~\cite{rudin2019stop}. To mitigate this, the field of Explainable AI (XAI) emerged with post-hoc and ante-hoc methods. Post-hoc methods try to use already trained backbone models and add an explainer on top to provide explanations, however they often provide fragile or misleading approximations of model behavior. That is why ante-hoc (self-explainable) methods were introduced. Within this paradigm, one of the most leading approaches is based prototypical part networks~\cite{chen2019looks} that offers transparency by utilizing similarity-based reasoning to identify representative latent patterns to justify model predictions.

\begin{figure}
    \centering
    \includegraphics[width=0.9\linewidth]{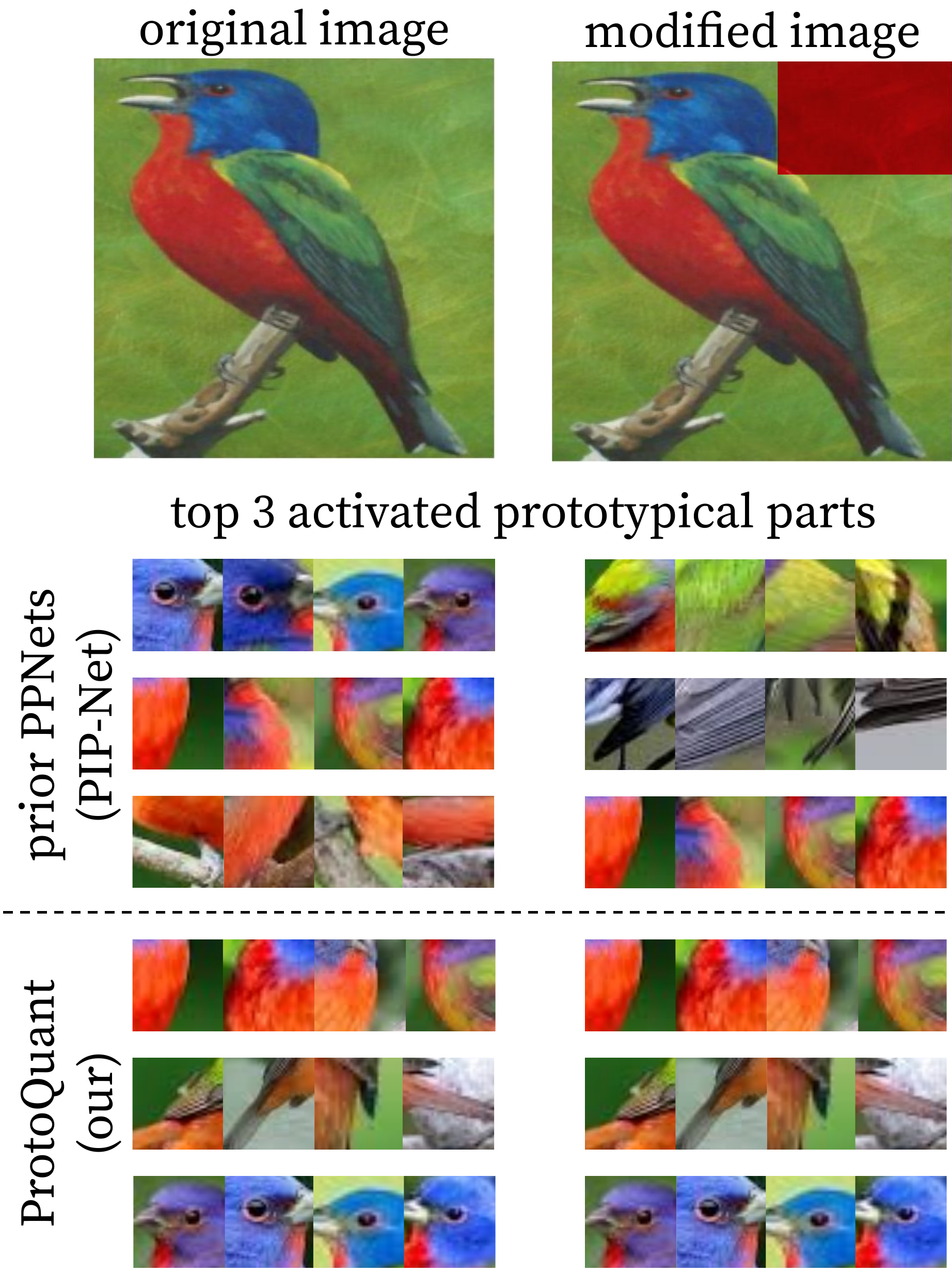}
    \caption{Standard ProtoPNet-based models (e.g. PIP-Net) exhibit representational instability: modifying a local region of an image can cause a drastic shift in the highest-activating prototypes, even if the modification is in a non-relevant area. Contrary, \our{} maintains stable activations despite input modifications.}
    \label{fig:teaser}
    \vspace{-1em}
\end{figure}

Despite their conceptual appeal, existing prototypical architectures face fundamental limitations regarding grounding, where prototypes may not corresponds to parts of training images, and representational stability (see Figure~\ref{fig:teaser}), as their prototypes activation may drastically change under small image perturbation as shown in~\cite{sacha2024interpretability}. Specifically, methods such as PIP-Net~\cite{nauta2023pip} and InfoDisent~\cite{struski2024infodisent} learn latent vectors that may not correspond to tangible, realizable training patches representing the visual concept. On the other hand, prior architectures such as~\cite{chen2019looks,nauta2021neural} often suffer from representational instability, where the model decision changes due to the shifts in prototype activation under minor input perturbations. This susceptibility to perturbations does not align with the principles of robust human perception, where we identify similarly the same objects in slightly different inputs. Additionally, prototypical models struggled with scalability to large-scale benchmarks such as ImageNet resulting in poor accuracy performance.

In this work, we present ProtoQuant, an interpretable framework designed for frozen foundation models. ProtoQuant addresses the dual challenges of grounding and representational instability through Vector-Quantized Variational Autoencoder
(VQ-VAE)~\cite{van2017neural} mechanism. By discretizing the latent space into a structured concept codebook via a  quantization mechanism, we enforce a rigorous mapping where every extracted feature vector is projected onto its nearest verifiable concept. This ensures that the model’s decision-making process is anchored in a finite, human-interpretable vocabulary of visual primitives.

Our approach results in a substantial improvement of  explanation reliability. Most notably, ProtoQuant achieves state of the art results in representational invariance on the Spatial Misalignment benchmark, reducing the Prototype Activation Change (PAC) by several orders of magnitude. Furthermore, by utilizing a discrete codebook over a frozen backbone, our method enables instant structural editability~\cite{donnelly2025rashomon}. This allows practitioners to perform model debugging or "concept pruning" by neutralization of specific codebook entries without the need for expensive computational retraining.

Empirically, ProtoQuant achieves competitive results on ImageNet compared to other prototypical parts-based methods. Evaluations on fine-grained datasets, including CUB-200 and Oxford Flowers, demonstrate that ProtoQuant maintains competitive predictive performance while providing superior stability of prototypes activations. Moreover, due to the lack of backbone tuning, \our{} is computationally efficient and can be easily applied to both, ViTs and CNNs. 

Our contributions are summarized as follows:
\begin{itemize}
    \item We introduce a quantization-based grounding mechanism that discretizes the feature space into a codebook of high stability concepts.

    \item We design ProtoQuant to work as an interpretable head for frozen backbones, allowing for seamless integration with modern backbone architectures without requiring full-parameter fine-tuning.

    \item We demonstrate that prototypical reasoning can scale to ImageNet with high fidelity, significantly bridging the trade-off between interpretability and predictive performance.

    \item We demonstrate state-of-the-art results on fine-grained datasets such as Oxford Flowers and CUB-200.

\end{itemize}

\begin{figure*}[t]
    \centering
    \includegraphics[width=1.0\textwidth]{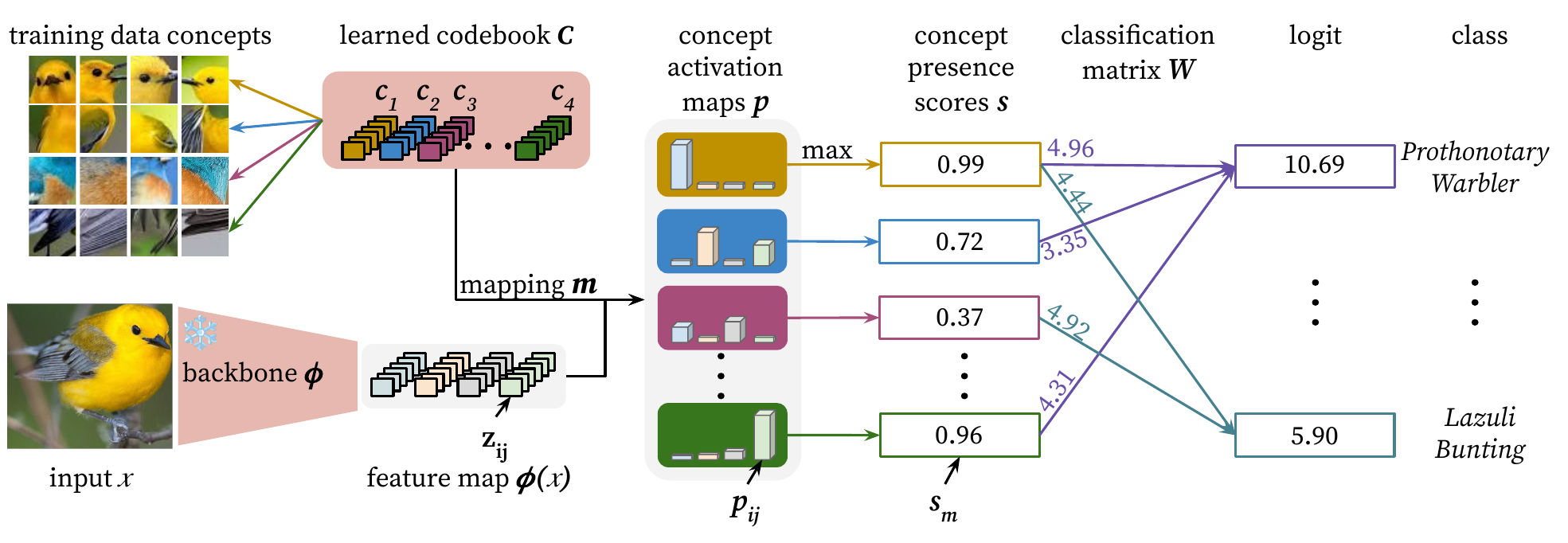}
    \caption{ Architecture of ProtoQuant. Our model consists of a frozen backbone $\phi$ that produces feature map $\phi(x)$ from input image $x$. Then each feature vector $z_{ij}$ is compared with prototypical parts from learned codebook $C$ through a mapping function $m$. As a result, we obtain concept activation maps $p$ over which we take the maximum value to obtain concept presence scores $s$, where $s_m$ corresponds to the probability of a presence of code $c_m$. Lastly, those score are multiplied with classification matrix $W$ to obtain final logit values.
    }
    \label{fig:method}
    \vspace{-1em}
\end{figure*}
\section{Related Works}

The pursuit of transparency in deep learning has led to a variety of eXplainable AI (XAI) strategies, ranging from post-hoc visualizations to inherently interpretable "ante-hoc" architectures~\cite{rudin2019stop}. Traditional attribution-based methods, such as Grad-CAM~\cite{selvaraju2017grad}, and Integrated Gradients~\cite{sundararajan2017axiomatic}, identify important pixels by analyzing network gradients of a trained model. To move beyond pixel-level importance, Concept Bottleneck Models (CBMs)~\cite{koh2020concept} constrain the network to predict high-level human concepts before reaching a final decision, while Concept Whitening~\cite{chen2020concept} and Concept Activation Vectors~\cite{kim2018interpretability} seek to align internal latent dimensions with predefined semantic categories. In contrast, case-based reasoning models, primarily Prototypical Parts Networks~\cite{chen2019looks}, incorporate interpretability into the decision process itself by identifying representative "prototypes" that justify a classification through a "this looks like that" comparison.

Prototypical part networks have evolved through various architectural refinements to enhance clarity and efficiency. The foundational ProtoPNet~\cite{chen2019looks} introduced the core concept of comparing image patches to learned exemplars in a transparent embedding space. TesNet~\cite{wang2021interpretable} and Deformable ProtoPNet~\cite{donnelly2022deformable} advanced this by enforcing orthogonality in prototype construction and allowing for spatial deformation to capture flexible object parts. ProtoPShare~\cite{rymarczyk2021protopshare}, ProtoTree~\cite{nauta2021neural}, and ProtoPool~\cite{rymarczyk2022interpretable} focus on efficiency by sharing prototypes across classes, organizing them into binary trees, or utilizing differentiable assignment to reduce the total prototype count. ProtoKNN~\cite{ukai2022looks} leverages $k$-nearest neighbor logic for similarity-based classification , while ProtoMIL~\cite{rymarczyk2022protomil} applies the paradigm to multiple instance learning. ProtoSeg~\cite{sacha2023protoseg} extends interpretability to semantic segmentation by assigning prototypical parts to individual pixels , and ICICLE~\cite{rymarczyk2023icicle} adapts the framework for interpretable continual learning. ProtoViT~\cite{ma2024interpretable} generalizes this idea to ViT models, while ProtoS-ViT~\cite{turbe2024protos} is applicable to any frozen ViT backbone. Recent methods such as PIP-Net~\cite{nauta2023pip, struski2024infodisent, wang2024mcpnet} utilize patch-based intuitive prototypes to enhance local reasoning but they lack prototypes grounding as they are based on continual or relative latent space. This rule is further followed by EPIC~\cite{borycki2025epic}, SIDE~\cite{dubovik2025side}, and other models. ProGReST~\cite{rymarczyk2023progrest} applies prototype logic to graph neural networks for molecular property prediction, while XProtoNet~\cite{kim2021xprotonet} provides global and local explanations for diagnosis in radiography. Finally, LucidPPN~\cite{pach2024lucidppn} addresses feature ambiguity by disentangling color from shape and texture, providing separate reasoning branches for granular explanations.

\section{ProtoQuant}

Our goal is to make the decision-making process of a frozen and pre-trained image classifier interpretable by grounding its predictions in a set of human-understandable \emph{concepts}. We achieve this by quantizing the model’s feature space into a discrete \emph{concept codebook}, replacing the classification head with a concept-based module as presented in Figure~\ref{fig:method}.

\subsection{Problem Setup}

Let $f: \mathbb{R}^{c \times h \times w} \rightarrow \mathbb{R}^{k}$ denote a pre-trained image classifier mapping an input image $\boldsymbol{x} \in \mathbb{R}^{c \times h \times w}$ to $k$ class logits. The network is decomposed into a \emph{backbone} $\phi: \mathbb{R}^{c \times h \times w} \rightarrow \mathbb{R}^{d \times H \times W}$, which produces a $d$-dimensional feature map, and a \emph{head} $h: \mathbb{R}^{d \times H \times W} \rightarrow \mathbb{R}^{k}$, typically implemented as global average pooling followed by a linear classifier:
\vspace{-0.5em}
\begin{equation}
    f(\boldsymbol{x}) = h(\phi(\boldsymbol{x})).
\end{equation}
Our objective is to transform this model into a \emph{concept-based classifier} that enables direct interpretation of predictions through a finite set of learned concepts.

\subsection{Concept Codebook Learning}
\label{subsec:codebook_learning}

To obtain a discrete representation of the embedding space, we learn a \emph{concept codebook} in a vector-quantized variational autoencoder (VQ-VAE) style, while keeping the backbone $\phi$ frozen.  

\noindent Let $\boldsymbol{z}_{ij} \in \mathbb{R}^{d}$ denote the feature vector at spatial location $(i,j)$ of $\phi(\boldsymbol{x})$. We maintain a codebook
\vspace{-0.5em}
\begin{equation}
    \mathcal{C} = \{\boldsymbol{c}_1, \dots, \boldsymbol{c}_m\}, \quad \boldsymbol{c}_i \in \mathbb{R}^{d},
\end{equation}
containing $m$ concept prototypes. Each feature vector is assigned to its nearest code based on cosine similarity:
\vspace{-0.5em}
\begin{equation}
    q(\boldsymbol{z}_{ij}) = \arg\max_{k} \bigl(\cos(\boldsymbol{z}_{ij}, \boldsymbol{c}_k) \bigr).
\end{equation}

\noindent We optimize the codebook vectors $\boldsymbol{c}_i$ by minimizing the \emph{codebook loss}:
\vspace{-0.5em}
\begin{equation}
    \mathcal{L}_{\text{codebook}} = 
    \frac{1}{N} \sum_{i,j} 
    \bigl\| \text{sg}[z_{ij}] - c_{q(z_{ij})} \bigr\|_2^{2},
\end{equation}

\noindent where $\text{sg}[\cdot]$ is the stop-gradient operator, preventing gradients from flowing into the embeddings $z_{ij}$.  
In other words, only the codebook vectors are updated during training, while the backbone parameters and their outputs remain fixed.  

This procedure discretizes the continuous embedding space into a finite \emph{concept vocabulary} that faithfully represents the feature distribution produced by a backbone $\phi$.






\subsection{Concept-Based Prediction Head}
\label{subsec:concept_head}
\noindent After obtaining the concept codebook, we replace the original head $h$ with a \emph{concept-based module} $\tilde{h}$ inspired by PIP-Net. The head is modularized into three key functions:

\paragraph{Concept Matching.}  
We denote the concept matching function by $m$:
\vspace{-0.5em}
\begin{equation}
m: (\boldsymbol{z}, \mathcal{C}, \alpha) \mapsto \boldsymbol{p},
\end{equation}

\noindent where $\boldsymbol{z} \in \mathbb{R}^{H \times W \times D}$ are spatial feature vectors, $\mathcal{C} = \{\boldsymbol{c}_1, \dots, \boldsymbol{c}_M\}$ is the concept codebook, and $\alpha$ is a temperature parameter.  
For each spatial feature vector $\boldsymbol{z}_{ij}$, we compute its cosine similarity with each concept and apply a sharp softmax:
\vspace{-0.5em}
\begin{equation}
p_{ij}(m) = 
\frac{\exp\!\bigl(\alpha \cdot \cos(\boldsymbol{z}_{ij}, \boldsymbol{c}_m)\bigr)}
{\sum_{m'=1}^{M} \exp\!\bigl(\alpha \cdot \cos(\boldsymbol{z}_{ij}, \boldsymbol{c}_{m'})\bigr)},
\end{equation}

\noindent resulting in concept probabilities $\boldsymbol{p} \in \mathbb{R}^{H \times W \times M}$.

\paragraph{Concept Activation Maps.}  
We denote the aggregation function by $a$:
\vspace{-0.5em}
\begin{equation}
a: \boldsymbol{p} \mapsto \boldsymbol{s},
\end{equation}

\noindent where $\boldsymbol{s} \in \mathbb{R}^{M}$ is a vector of concept presence scores.  
Each score is obtained by spatial max-pooling:
\vspace{-0.5em}
\begin{equation}
s_m = \max_{i,j} p_{ij}(m).
\end{equation}

\paragraph{Concept-to-Class Mapping.}  
We denote the class mapping function by $c$:
\vspace{-0.5em}
\begin{equation}
c: (\boldsymbol{s}, \boldsymbol{W}) \mapsto \hat{\boldsymbol{y}},
\end{equation}

\noindent where $\boldsymbol{W} \in \mathbb{R}_{+}^{k \times M}$ is a non-negative classification matrix, and the class logits are computed as
\vspace{-0.5em}
\begin{equation}
\hat{\boldsymbol{y}} = \boldsymbol{W} \boldsymbol{s}.
\end{equation}

\begin{figure}
    \centering
    \includegraphics[width=\linewidth]{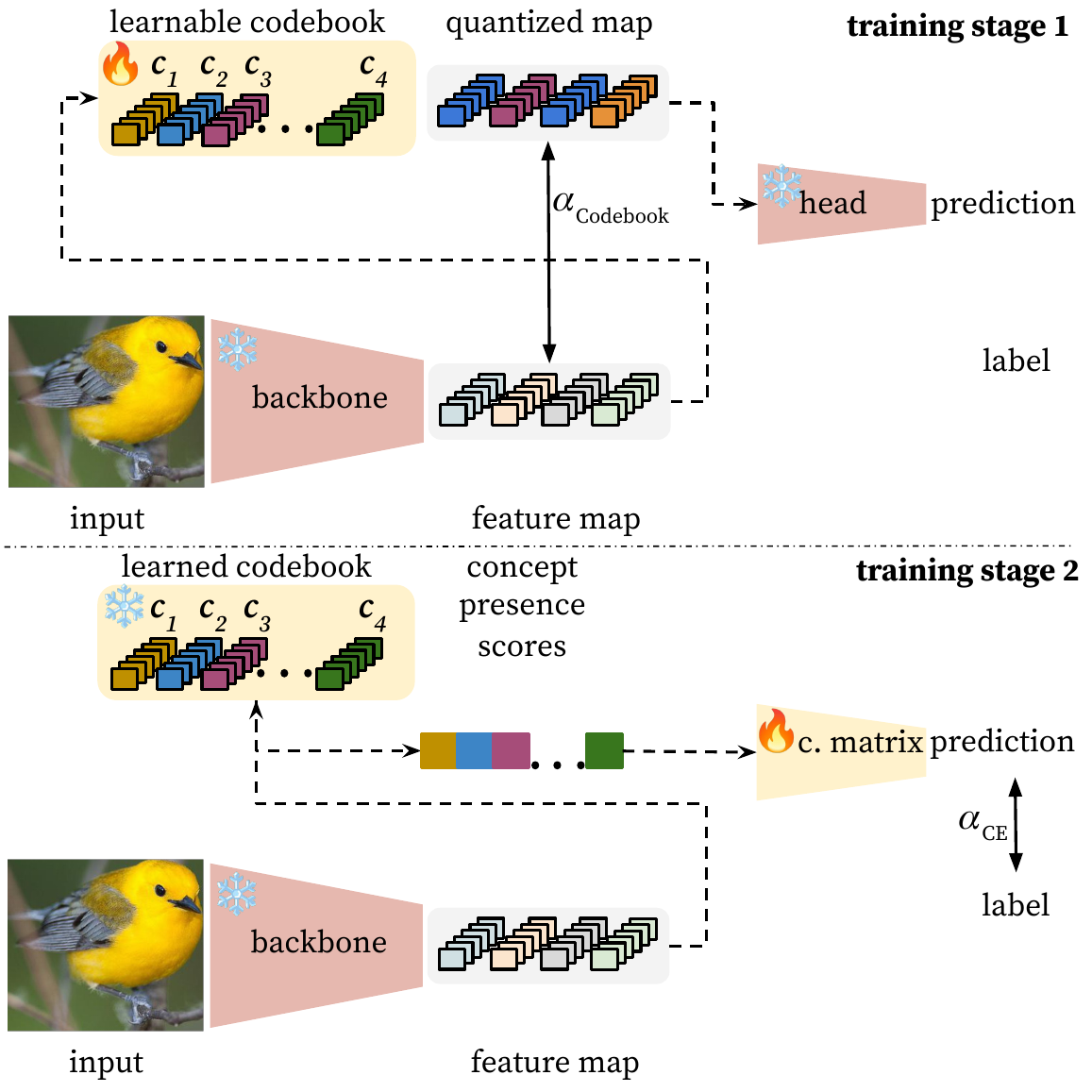}
    \caption{Training of \our{} is in two stages. The first one focuses on training a codebook to obtain meaningful prototypical parts, while the second switches original head with sparse and non-negative classification matrix to ensure interpretability.}
    \label{fig:training}
    \vspace{-1em}
\end{figure}

\paragraph{Interpretable Head.}  
The modular concept-based head can then be written as a composition of these functions:

\vspace{-0.5em}
\begin{equation}
\tilde{h}(\boldsymbol{z}) = \big(c \circ a \circ m\big)(\boldsymbol{z}, \mathcal{C}, \alpha, \boldsymbol{W}).
\end{equation}

\noindent This modular formulation clarifies the role of each component and also enables flexible analysis and visualization of how individual concepts contribute to the final predictions.



\subsection{Training}

\our{} utilizes a two-stage training regime (see Figure~\ref{fig:training}) to ensure representational stability of prototypes.

\textbf{Stage 1: Unsupervised Grounding:} The pre-trained backbone is frozen. We initialize the codebook with $K$ codes (chosen per dataset, with $10$ times the number of classes as rule of thumb) and train it as detailed in Section~\ref{subsec:codebook_learning} in an unsupervised way. We leverage the frozen pretrained head only for validation of the codebook in terms of validation accuracy, but it's not used in codebook optimization. To ensure arhcitectural agnosticism we adapt spatial features: for CNNs we flatten the final feature map into spatial-channel matrix, while for Vision Transformers, we discard the \texttt{[CLS]} token and utilize patch tokens to preserve localized part information. Since standard ViT backbones are pre-trained for \texttt{[CLS]}-based classification, our prediction head aggregates these patch embeddings via Global Average Pooling (GAP) to align the frozen features with the requirements of discriminative classification.

\textbf{Stage 2: Supervised Reasoning:} The pretrained head is replaced with an interpretable head described in Section~\ref{subsec:concept_head}, the backbone is frozen and we use the codebook from Stage 1. In this stage we use a standard cross entropy loss to train and validate the head. After this stage we may prune the weights to top $K$ per class and remove corresponding codes, increasing the interpretability and reducing the codebook.

\paragraph{Summary.}Our method transforms a black-box classifier into a concept-based reasoning model that (i) learns a discrete set of prototypical features, (ii) predicts class scores as a non-negative combination of concept activations, and (iii) visualizes which regions of the image are responsible for each concept. This pipeline enables both \emph{global explanations} (via concept-to-class weights) and \emph{local explanations} (via concept activation maps). Moreover, the training process is simplified, and requires only two stages.


\vspace{-0.5em}
\section{Experimental Setup}
\label{sec:experimental_setup}

In this section we provide details on the experimental setup of our work. We make the code available (in supplement)\footnote{Code will be released as public repository in a camera-ready version of the work.}.

\paragraph{Datasets.}
We evaluate our method on four standard fine-grained benchmarks: CUB-200-2011 (CUB-200)~\cite{wah2011caltech}, containing 200 bird species; Stanford Cars (CARS)~\cite{krause20133d}, with 196 car models; Stanford Dogs (DOGS)~\cite{khosla2011novel}, comprising 120 dog breeds; and Oxford 102 Flowers (FLOWERS)~\cite{nilsback2008automated}, featuring 102 flower categories. To demonstrate scalability, a key contribution of our work, we also report results on the large-scale ImageNet-1K dataset~\cite{deng2009imagenet}. Finally, we utilize the FunnyBirds~\cite{hesse2023funnybirds} synthetic dataset and Spatial Misalignament Benchmark~\cite{sacha2024interpretability} for evaluation of part-based interpretability.

\paragraph{Architecture.}
For Accuracy, Purity~\cite{nauta2023pip} and Spatial Misalignment experiments we test various popular backbones including CNN's like ResNet-50~\cite{he2016deep}, ConvNeXt-Tiny/Large~\cite{liu2022convnet}, and the DeiT-Small ViT~\cite{touvron2022deit}, all pretrained on ImageNet. In case of FunnyBirds benchmark we use the ViT-B-16 and ResNet50 provided by the authors.

Details on training parameters, evaluation metrics, setup specific for FunnyBirds can be found in the Appendix~\ref{sec:exp_set_app}. 

\begin{table*}[t]
    \centering
    \caption{\our{} consistently outperforms other prototypical parts models across the majority of metrics. Notably, the superior scores in PRC and PAC (in ViT) demonstrate high representational stability and minimal activation variance, ensuring that predictive accuracy remains robust (low AC). While \our{} achieves comparable value of PLC metric, which indicate a shift in activation localization, we attribute this to the inaccuracy of current prototype activation visualization methodologies.}
    \label{tab:spatial_misalignment_table}
    \small
    \begin{tabular}{lcccccc}
        \toprule
         \textbf{Model} & \textbf{PAC} $\downarrow$ & \textbf{PLC} $\downarrow$ & \textbf{PRC} $\downarrow$ & \textbf{Acc. Before} $\uparrow$ & \textbf{Acc. After} $\uparrow$ & \textbf{AC} $\downarrow$ \\
        \midrule
        \multicolumn{7}{c}{\textit{CNN}} \\
        \midrule
        ProtoPNet  & 23.70 & 24.00 & 13.50 & 76.40 & 68.20 & 8.20 \\
        TesNet     & \textbf{3.40} & 16.00 & 2.90 & 81.60 & 75.80 & 5.80 \\
        ProtoPool  & 11.20 & 31.80 & 4.50 & 80.80 & 76.00 & 4.80 \\
        ProtoTree  & 23.70 & 27.70 & 13.50 & 76.40 & 68.20 & 8.20 \\
        \our{} & 6.80 & \textbf{11.72} & \textbf{0.34} & \textbf{87.92} & \textbf{86.19} & \textbf{1.73} \\
        \midrule
        \multicolumn{7}{c}{\textit{ViT}} \\
        \midrule
        InfoDisent & 0.11 & \textbf{0.37} & 13.66 & 83.66 & 67.21 & 16.45 \\
        ProtoViT   & 2.92 & 21.68 & 1.28 & \textbf{85.40} & 82.80 & 2.60 \\
        \our{} & \textbf{0.07} & 20.30 & \textbf{0.23} & 85.21 & \textbf{84.86} & \textbf{0.35} \\
        \bottomrule
    \end{tabular}
\end{table*}


\begin{table*}[t]
\caption{Top-1 Classification Accuracy on fine-grained benchmarks.  Our method demonstrates competitive or superior performance, particularly with the ConvNeXt backbone. 
}
\label{tab:datasets_accuracy_results}
\centering
\begin{small}
\begin{sc}
\begin{tabular}{lcccc}
\toprule
\textbf{Method} & \textbf{CUB-200} & \textbf{Cars} & \textbf{Dogs} & \textbf{Flowers} \\
\midrule
\textit{ResNet-50} & $84.5$ & $89.0$ & $89.0$ & $92.8$ \\
\cmidrule{2-5}
ProtoPool     & $85.5$ & $91.6$ & -- & -- \\
Q-SENN         & $\mathbf{85.9}$ & $\mathbf{92.9}$ & -- & -- \\
PIP-Net       & $82.0${\scriptsize$\pm0.3$} & $86.5${\scriptsize$\pm0.3$} & -- &  -- \\
ProtoArgNet   & $85.4$ & $89.3$ & -- & -- \\
InfoDisent    & $83.0$ & $\mathbf{92.9}$ & $86.6$ & -- \\
\our{} (Ours) & $82.5$\scriptsize$\pm 0.1$ & $88.2$\scriptsize$\pm 0.1$ & $\mathbf{88.0}$\scriptsize$\pm 0.2$ & $\mathbf{95.4}$\scriptsize$\pm 0,1$ \\
\midrule
\textit{ConvNeXt-Tiny} & $88.4$ & $92.4$ & $92.0$ & $97.1$ \\
\cmidrule{2-5}
Lucid PPN     & $81.5${\scriptsize$\pm0.4$} & $91.6${\scriptsize$\pm0.2$} & $79.4${\scriptsize$\pm0.4$} & $95.0${\scriptsize$\pm0.3$} \\
PIP-Net       & $84.3${\scriptsize$\pm0.2$} & $88.2${\scriptsize$\pm0.5$} & $80.8${\scriptsize$\pm0.4$} & $91.8${\scriptsize$\pm0.5$} \\
InfoDisent    & $84.1$ & $90.2$ & -- & -- \\
MCPNet        & $83.5$ & -- & -- & -- \\
\our{} (Ours) & $\mathbf{87.6}$\scriptsize$\pm0.1$  & $\mathbf{92.6}$\scriptsize$\pm0.2$ & $\mathbf{91.2}$\scriptsize$\pm0.1$ & $\mathbf{97.0}$\scriptsize$\pm0.1$ \\
\midrule
\textit{DeiT-Small} & $86.9$ & $89.9$ & $89.8$ & $96.4$ \\
\cmidrule{2-5}
ProtoViT      & $85.4$ & $\mathbf{91.8}$ & -- & -- \\
ProtoPFormer  & $84.9$ & $90.1$ & $\mathbf{90.0}$ & -- \\
\our{} (Ours) & $\mathbf{86.9}${\scriptsize$\pm0.2$} & $91.0${\scriptsize$\pm0.2$} & $89.8${\scriptsize$\pm0.2$} & $\mathbf{95.01}${\scriptsize$\pm0.1$} \\
\bottomrule
\end{tabular}
\end{sc}
\end{small}
\end{table*}

\begin{figure}[t]
\includegraphics[width=0.5\textwidth]{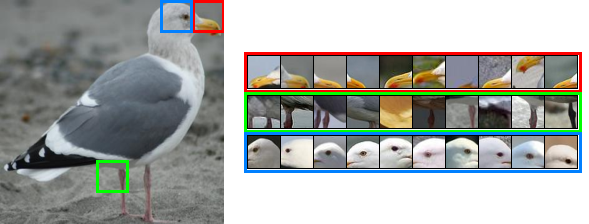}
    \caption{Visualization of the top 3 activated prototypes from \our{} on an example from CUB-200.}    \label{fig:gull_example}
    \vspace{-1em}
\end{figure}

\section{Results and Analysis}
\label{sec:results}

Here, we focus on the performance of \our{}. Firstly, we investigate how \our{} is susceptible to representational instabilities on Spatial Misalignement Benchmark, then we compare its accuracy with other prototypical-parts based methods on fine-grained vision datasets and how well it scales to ImageNet. Then, we analyze the purity of prototypes for our method together with explanations quality on FunnyBirds framework (see Appendix~\ref{sec:fb}). Furthermore, we perform ablations to show how codebook size impacts the model performance, then how enforcing sparsity of explanations impacts its effectiveness. Additionally, we provide the analysis on computational resources needed to train \our. Lastly, we provide visualization (see Figure~\ref{fig:gull_example}) of explanations generated by \our{}, and more examples can be found in Appendix~\ref{sec:vis}.

\paragraph{Representational Instability via Spatial Misalignment.}

One of the most critical limitations of prototypical-parts network is their representational instability. Specifically, when input image is slightly modified, a different set of prototypical parts may activate, completely changing the model's behavior and decisions. Metrics such as PAC and PRC from Spatial Misalignement Benchmark are dedicated to measure this property, and in Table~\ref{tab:spatial_misalignment_table} we report comparison of two different backbone of \our{} to other models.

Across both backbones, \our{} is clearly superior in the metrics measuring if the model's behavior changed after input perturbations: Accuracy Change (AC) and Rank Change (PRC). Our ViT variant achieves a nearly perfect AC of 0.35, while our ConvNeXt-Tiny variant reaches 1.73. This consistency suggests that our method preserves the pre-trained backbone's inherent ability to filter background noise, with the quantization head resulting in representational stability. 

In terms of Prototypical Activation Change (PAC), we observe significant differences between backbones (0.07 for DeiT-S vs 6.80 for Conv-T), likely reflecting the Transformer's global attention stability compared to the local receptive fields of CNNs. For Location Change (PLC), \our{} is highly competitive with standard part-prototype methods. While InfoDisent achieves an outlier result in this category (PLC = 0.37), all other evaluated methods exhibit significantly higher values competetive to \our{}; the mechanism behind InfoDisent's spatial stability may be related to disentanglement process. Overall, \our{} is either superior or highly competitive across all evaluated metrics, providing the most robust and logically consistent grounding and representational stability among current prototypical-part methods.

\paragraph{Classification Accuracy.}

\begin{table}[t]
\caption{ImageNet-1k Accuracy. Comparison against the black-box baseline (reported in parentheses) and the InfoDisent method. Our method achieves superior results over InfoDisent, reducing the gap of ante-hoc methods to non-interpretable baselines.
*ViT-B-16 is finetuned using GAP head instead of a CLS token, CLS version has 81.07 accuracy.}
\label{tab:imagenet_results}
\centering
\begin{small}
\begin{sc}
\begin{tabular}{lc}
\toprule
\textbf{Method} & \textbf{Top-1 Accuracy} \\
\midrule
\multicolumn{2}{l}{\textit{ResNet-50 (Baseline: 80.85)}} \\
InfoDisent & $67.80$ \\
\our{} (Ours)       & $\mathbf{77.86}${\scriptsize$\pm 0.3$} \\
\midrule
\multicolumn{2}{l}{\textit{ConvNeXt-Large (Baseline: 84.14)}} \\
InfoDisent & $82.80$ \\
\our{} (Ours)       & $\mathbf{82.87}${\scriptsize$\pm 0.3$} \\
\midrule
\multicolumn{2}{l}{\textit{ViT-B/16 (Baseline: 80.70*)}} \\
InfoDisent & $79.20$ \\
\our{} (Ours)       & $\mathbf{80.13}${\scriptsize$\pm 0.2$} \\
\bottomrule
\end{tabular}
\end{sc}
\end{small}
\vspace{-0.5em}
\end{table}

Table~\ref{tab:datasets_accuracy_results} summarizes the results on fine-grained benchmarks. \our{} consistently achieves or exceeds the performance of state-of-the-art interpretable models within the same backbone constraints, outperforming even method which finetune the backbone, such as PIP-Net. Notably, on the CUB-200 dataset using a ConvNeXt-Tiny backbone, \our{} reaches highest accuracy among the methods, outperforming PIP-Net ($84.3\%$) and MCPNet ($83.45\%$). Since \our{} operates on a frozen backbone, its effectiveness is intrinsically tied to the quality of the underlying backbone; the more semantically structured the backbone's latent space, the more efficiently our codebook can discretize discriminative concepts. 

A key contribution of our work is scalability. As shown in Table~\ref{tab:imagenet_results}, \our{} is one of the few prototypical-parts methods able to handle ImageNet. Unlike InfoDisent, which suffers a $\sim13\%$ drop on ResNet-50 compared to the black-box baseline, \our{} maintains a much tighter margin ({77.86\%} vs $80.85\%$). On ConvNeXt-Large, \our{} achieves {82.87\%}, demonstrating that quantization-based interpretability can scale to high-capacity architectures where the "representational ceiling" of the backbone is higher.

\paragraph{Semantic Purity and Grounding.}
We evaluate the semantic quality of learned concepts using the Purity metric (Table~\ref{tab:purity_results}), which assesses how consistently prototypes correspond to human-annotated parts. Our results indicate that \our{} is competitive with or superior to nearly all existing methods evaluated under similar architectural constraints. While the original PIP-Net~\cite{nauta2023pip} reports much higher purity score ($0.93$), it modifies the backbone strides to produce significantly larger ($28 \times 28$) feature maps, and finetunes the backbone to adjust for this.

\begin{table}[t]
  \caption{\our{} achieves highest purity of CUB-prototypes w.r.t. object part annotations when compared to other prototypical-parts method except PIP-Net which performs self-supervision pretraining to align representations. However, original PIP-Net calculated this metric for a layer with different stride. When unifying the last layer and freezing the backbone, PIPNet obtains much lower scores (see PIPNet*).}
  \label{tab:purity_results}
  \centering
  \begin{small} 
  \begin{sc}    
  \begin{tabular}{@{}lcc@{}}
    \toprule
    \textbf{Method} & \textbf{Purity train} & \textbf{Purity test} \\
    \midrule
    \multicolumn{3}{c}{Fine-tuned backbone}\\
    \midrule
    ProtoPNet & $0.44 \pm 0.21$ & $0.46 \pm 0.22$ \\
    ProtoPool & $0.35 \pm 0.20$ & $0.36 \pm 0.21$\\
    ProtoTree & $0.13 \pm 0.14$ & $0.14 \pm 0.16$\\
    ProtoPShare & $0.43 \pm 0.21$ & $0.43 \pm 0.22$\\
    PIPNet & $\mathbf{0.92 \pm 0.16}$ & $\mathbf{0.93 \pm 0.15}$\\
    \midrule
    \multicolumn{3}{c}{Frozen backbone}\\
    \midrule
    PIPNet* & $0.35 \pm 0.22$ & $0.34 \pm 0.23$ \\
    \our{} (Ours) & $\mathbf{0.47 \pm 0.26}$ & $\mathbf{0.47 \pm 0.25}$ \\
    \bottomrule
  \end{tabular}
  \end{sc}
  \end{small}
\vspace{-1.25em}
\end{table}

To fairly evaluate \our{}, we test against a modified baseline (PIP-Net\textsuperscript{*}) that uses standard $7 \times 7$ feature maps. In this setting, we unfroze the final stage of the backbone for PIP-Net\textsuperscript{*} to allow its prototypes to adapt. In this comparison \our{} outperforms the PiP-Net (0.35). Notably, \our{} is also competitive with methods like ProtoPNet ($0.46$), despite the fact that \our{} maintains a frozen backbone. Although the relatively high standard deviation reflects the inconsistence in concepts derivation, these results suggest that our quantization mechanism effectively anchors prototypes to meaningful visual concepts without requiring the specialized architectural modifications or full-backbone tuning used by prior work.

\begin{figure}[t]
\includegraphics[width=\linewidth]{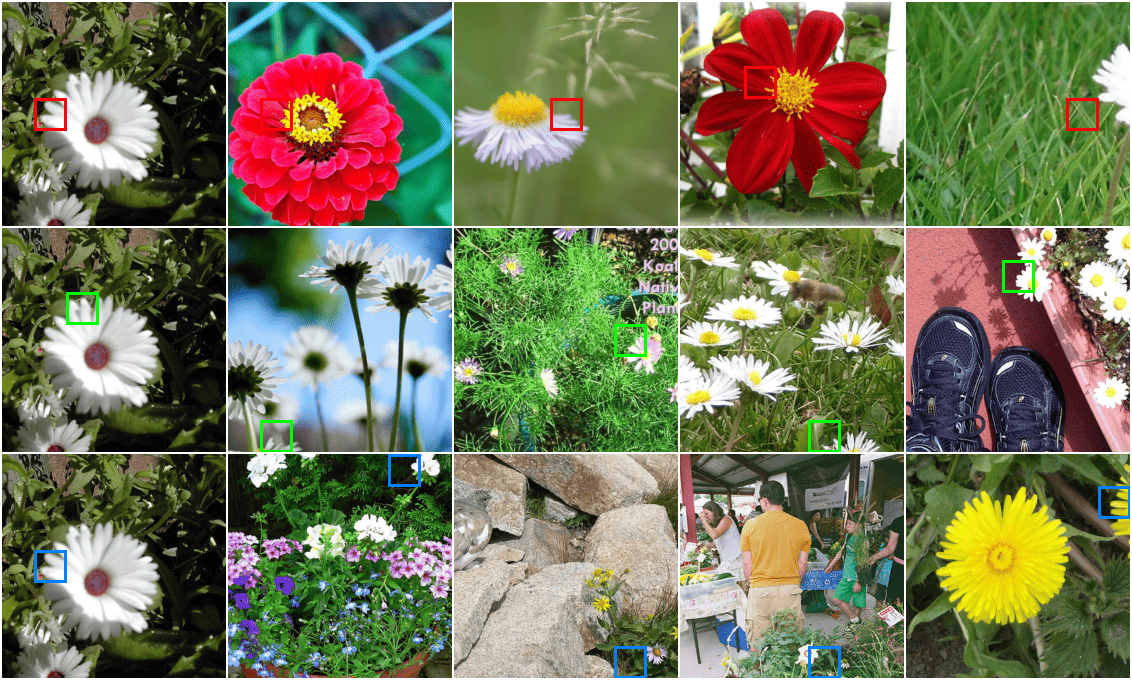}
\caption{Representative prototypical explanations on ImageNet. The input (daisy) is shown in the first column, with rows visualizing the most important prototypical parts via their nearest neighbors in the training set (columns 2–5). The activation of patches from multiple distinct classes for a single prototype highlights the capacity of \our{} for inter-class concept sharing, where visual primitives are leveraged as a universal basis for classification.}
\label{fig:cub_multi}
\vspace{-1em}
\end{figure}

\begin{figure}
    \centering
    \includegraphics[width=\columnwidth]{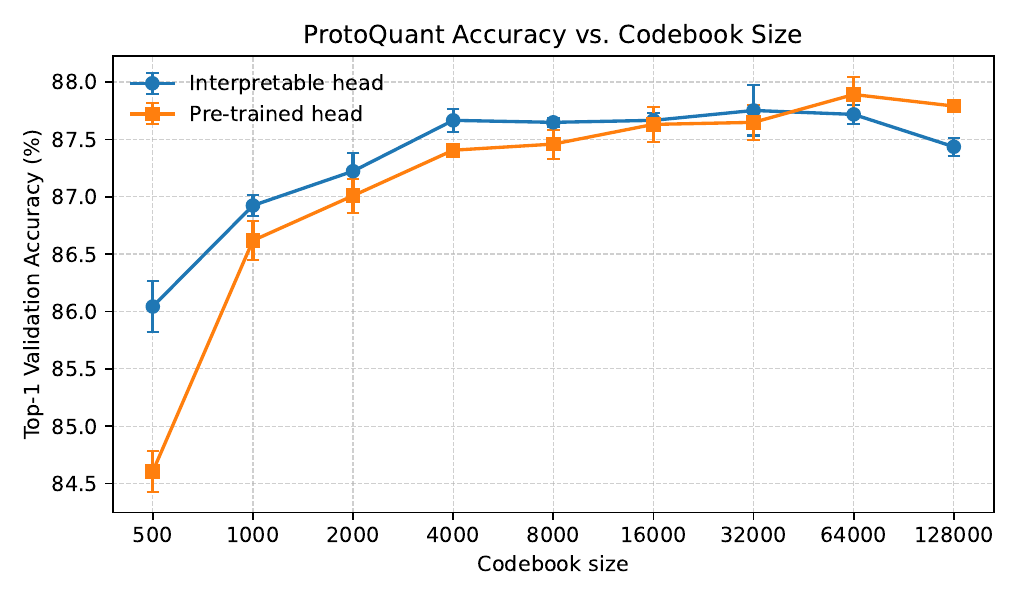}
    \caption{Top-1 accuracy on CUB-200 (ConvNeXt-Tiny) comparing the pre-trained head (Stage 1) and our interpretable head (Stage 2). The plot showcase that the increase in codebook size provides gain in accuracy however, it saturates with a codebook size 32000. Moreover, the interpretable head does not hurt model's accuracy. Error bars are standard deviation of three runs.}
\label{fig:codebook_size_ablation}
\vspace{-0.5em}
\end{figure}

\paragraph{Impact of Codebook Capacity.}
We analyze the interaction between discretization levels and predictive performance using a ConvNeXt-Tiny backbone on the CUB-200 dataset. Following the two-stage protocol described in Section~\ref{sec:experimental_setup}, we evaluate the accuracy of the pre-trained head on quantized features versus our interpretable head trained on the frozen codebook. Results are summarized in Figure~\ref{fig:codebook_size_ablation}.

As expected, accuracy initially increases with the codebook size, as a larger vocabulary allows the discrete prototypes to more faithfully approximate the backbone's continuous latent space. \our{} demonstrates a significant regularizing effect at lower capacities; with a codebook size of only $500$, the interpretable head outperforms the pre-trained head by {+1.44\%}. This suggests that the discrete bottleneck effectively filters out non-discriminative latent noise from the ConvNeXt features.

Interestingly, performance remains remarkably stable across a wide range of capacities. While we observe a minor accuracy peak at $64,000$ codes, likely due to the model having a larger pool of potential prototypes to utilize, the results are highly similar to those achieved with significantly fewer codes. For instance, the accuracy at $128,000$ codes ($87.44\%$) is comparable to that of $4,000$ codes ($87.67\%$). This saturation suggests that beyond a certain point, additional prototypes provide diminishing returns and may eventually capture sample-specific artifacts rather than generalizable semantic parts. 

\paragraph{Sparsity and Representational Efficiency.}
We analyze the trade-off between explanation conciseness and predictive performance by restricting each class to its top-$K$ highest-weighted prototypes. Table~\ref{tab:combined_sparsity} illustrates the behavior of the Stage 2 (Frozen) model compared to the Pruned and Finetuned variant using a ConvNeXt-Tiny backbone. 

A substantial difference in representational efficiency is observed. The Stage 2 model achieves a high peak accuracy of {87.76\%} when utilizing the full codebook, but its performance degrades rapidly under sparsity constraints, falling to $67.21\%$ at $K=10$. This suggests that in a frozen latent space, discriminative information is spread across a large number of prototypes, requiring a significant amount of features to achieve high-fidelity classification. 

In contrast, the finetuning process acts as a semantic distiller. The Pruned and Finetuned model successfully concentrates the backbone's discriminative power into a highly compact vocabulary, maintaining its slightly lower accuracy (86.07\%) even when limited to only 10 prototypes per class. While the Stage 2 model has a slightly higher absolute ceiling, the finetuned variant is vastly more practical for interpretability.

\paragraph{Explanation visualizations.} Figure~\ref{fig:cub_vis_prototype2161}, Figure~\ref{fig:cub_multi} provides visualization of most activated prototypes for a given input image from CUB and ImageNet datasets, respectively. Our approach allows visualizations in a style of PIPNet as well as a grid of concepts on training images more similar to ProtoPNet style. In Appendix~\ref{sec:vis} we provide more examples.

\begin{table}[t]
\caption{Comparison of training times between full model (unfrozen backbone) and \our{} (frozen backbone). Working with frozen backbone reduces epoch time almost by half. Note that, it was calculated on CUB200 with ConvNeXt-T backbone and is averaged over 30 epochs.}
\label{tab:training_times}
\vspace{-1ex}
\begin{center}
\begin{small}
\begin{sc}
\begin{tabular}{lc}
\toprule
Training Type & Avg. Epoch Time (s) \\
\midrule
Full model (Unfrozen B.) & 22.033 $\pm$ 0.135 \\
\our~(Frozen B.) & 11.537 $\pm$ 0.020 \\
\midrule
\textbf{Average Time Saved} & \textbf{10.496 (47.6\%)} \\
\bottomrule
\end{tabular}
\end{sc}
\end{small}
\end{center}
\vspace{-2em}
\end{table}








\begin{figure}[t]
\centering
\begin{tikzpicture}
    \begin{axis}[
        width=0.48\textwidth,   
        height=5.5cm,
        xlabel={Prototypes per Class ($K$)},
        ylabel={Top-1 Accuracy (\%)},
        xmin=0, xmax=52,
        ymin=30, ymax=90,
        xtick={1, 10, 20, 30, 40, 50},
        grid=major,
        grid style={dashed, gray!30},
        legend pos=south east,
        legend style={nodes={scale=0.8, transform shape}}, 
        /pgf/number format/.cd,
        use comma,
        1000 sep={}
    ]

    \addplot[
        color=myblue,
        mark=*,
        thick,
        mark options={scale=0.8}
    ]
    coordinates {
        (1, 31.26)
        (5, 56.14)
        (8, 62.72)
        (10, 67.21)
        (20, 77.48)
        (30, 82.78)
        (50, 85.12)
    };
    \addlegendentry{Frozen Codes}

    \addplot[
        color=myred,
        mark=square*,
        thick,
        mark options={scale=0.8}
    ]
    coordinates {
        (1, 42.03)
        (5, 78.24)
        (8, 84.78)
        (10, 86.07)
    };
    \addlegendentry{Pruned \& Tuned}

    \addplot[
        color=mygray,
        dashed,
        domain=0:52,
        samples=2
    ] {87.76};
    \node[anchor=south east, color=mygray, font=\scriptsize] at (axis cs:50,88) {Full Baseline (87.8\%)};

    \end{axis}
\end{tikzpicture}
\caption{Comparison of Top-1 accuracy on CUB-200 as we restrict the number of prototypes allowed per class ($K$). While the \textit{Frozen} Stage 2 model has higher peak capacity, the \textit{Finetuned} variant is significantly more efficient, maintaining full accuracy with only 10 prototypes per class showcasing that tuning of codebook after Stage 1 of training may be beneficial. }
\label{fig:combined_sparsity}
\vspace{-1em}
\end{figure}
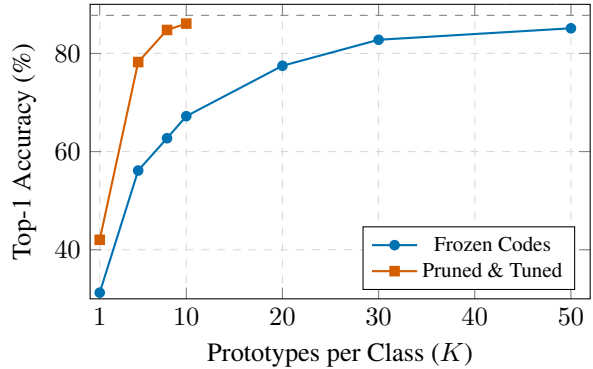

\paragraph{Computational efficiency.} Table~\ref{tab:training_times} shows a comparison on training time per epoch when we have a full model training, including backbone finetuning, and \our{}. The results shows that frozen backbone can allow for almost 50\% savings in training time.


\section{Conclusions}
In this work, we introduced ProtoQuant, a quantization-based framework that achieves grounded interpretability by discretizing the feature space into a concept codebook. Unlike prior methods, ProtoQuant ensures prototypes are faithful representations of training data without requiring backbone tuning, allowing it to scale effectively to large-scale benchmarks like ImageNet. Our results demonstrate that this modular approach maintains competitive accuracy across general and fine-grained tasks while providing robust, pixel-level explanations.

Despite its strengths, the reliance on a discrete codebook may limit representation flexibility compared to continuous spaces. Future work could focus on automating codebook size selection and exploring advanced pruning to further optimize the concept-to-class mapping. 

\paragraph{Impact Statement.}
This research advances Explainable AI (XAI) by providing a method to audit high-capacity models without sacrificing predictive power. By grounding model decisions in a verifiable vocabulary of visual concepts, ProtoQuant addresses "prototype drift", ensuring that explanations are based on actual data rather than abstract latent artifacts.

In high-stakes sectors such as medical imaging and autonomous driving, the grounded "this looks like that" reasoning allows practitioners to verify that a model is attending to clinically or operationally relevant features. Lastly, this work supports the deployment of more transparent, accountable, and trustworthy machine learning systems in critical real-world applications.

\section*{Acknowledgments}

The work was funded by the "Interpretable and Interactive Multimodal Retrieval in Drug Discovery" project. The "Interpretable and Interactive Multimodal Retrieval in Drug Discovery" project (FENG.02.02-IP.05-0040/23) is carried out within the First Team programme of the Foundation for Polish Science co-financed by the European Union under the European Funds for Smart Economy 2021-2027 (FENG). 

Some experiments were performed on servers purchased with funds from the flagship project entitled ``Artificial Intelligence Computing Center Core Facility'' from the DigiWorld Priority Research Area within the Excellence Initiative -- Research University program at Jagiellonian University in Kraków.

We gratefully acknowledge Polish high-performance computing infrastructure PLGrid (HPC Center: ACK Cyfronet AGH) for providing computer facilities and support within computational grant no. PLG/2025/018272




\bibliography{main}
\bibliographystyle{icml2026}

\newpage
\appendix
\onecolumn

\section{Additional Details on Experimental Setup}
\label{sec:exp_set_app}

\subsection{Training Details.}
All models are trained using standard ConvNeXt data augmentation pipelines from Torchvision, including random resized cropping to $224 \times 224$, random horizontal flipping, TrivialAugmentWide, MixUp, CutMix, and ImageNet normalization. We apply label smoothing with a factor of $0.1$.
Optimization is performed using AdamW~\cite{loshchilov2017decoupled} with a batch size of $1024$. Unless stated otherwise, the same training hyperparameters are used for both training stages. Each stage is trained for $30$ epochs, and the checkpoint with the highest validation accuracy is selected for evaluation.

\textbf{Optimization and Learning Rate Schedule.}
We use a base learning rate of $0.05$ with a weight decay of $5 \times 10^{-4}$, selected via a hyperparameter sweep. The learning rate schedule follows a cosine decay implemented in the \texttt{timm}~\cite{rw2019timm} library, with a $5$-epoch linear warm-up phase and a minimum learning rate of $1 \times 10^{-6}$. The relatively high base learning rate is stabilized by the large batch size and warm-up strategy.

\textbf{Codebook Initialization and Dynamics.}
To mitigate representational redundancy and prevent prototype collapse, the concept codebook $\mathcal{C}$ is initialized using an orthogonal initialization scheme. This initialization encourages prototypes to be well-separated in the latent space at the start of training. No explicit orthogonality constraint is enforced during optimization beyond initialization.

\textbf{Hardware and Infrastructure.}
All experiments are conducted on NVIDIA GH200 GPUs with $96$~GB memory. For smaller-scale datasets (CUB-200-2011 and Stanford Cars), training is performed on a single GPU. ImageNet-1K experiments are distributed across $4 \times$ GH200 GPUs to support the larger batch size and memory requirements.

\textbf{Temperature Parameter (\texorpdfstring{$\alpha$}{alpha}).}
The temperature parameter $\alpha$ controls the sharpness of the softmax used to pool prototype activations. Smaller values of $\alpha$ induce sharper competition between prototypes, encouraging each spatial location to activate only a small number of highly similar concepts. This leads to more localized and semantically distinct prototype activations, which empirically improves interpretability. Larger values of $\alpha$ result in smoother assignments and increased prototype co-activation.
Unless stated otherwise, we fix $\alpha = 0.1$ in all experiments.

\paragraph{Prototype Pruning.}
To improve interpretability and reduce redundancy among learned prototypes, we optionally apply classifier-driven prototype pruning after training.
Pruning is performed \emph{post hoc} and does not alter the training procedure or learned feature representations.

\textbf{Logical Pruning (Top-$k$ Masking).}
Given the non-negative classifier weight matrix $\boldsymbol{W} \in \mathbb{R}_{+}^{C \times M}$, we retain only the top-$k$ highest-weighted prototypes per class and mask all remaining weights to zero.
This limits the number of prototypes contributing to each class prediction while preserving the original codebook and backbone.
Logical pruning is used for controlled interpretability analyses without modifying the underlying model capacity.

\textbf{Physical Pruning (Codebook Shrinking).}
In addition, we support physical removal of inactive prototypes.
A prototype is considered active if it has a strictly positive classifier weight for at least one class.
Prototypes that are unused by all classes are removed from both the codebook and the classifier weight matrix, yielding a smaller model with identical predictions.
Fine-tuning can be performed after physical pruning.

Unless explicitly stated, pruning is not applied during training and is only used for interpretability analysis and model compression studies.

\subsection{Evaluation Metrics.}
We report top-1 classification accuracy on validation subset of the datasets as the primary performance metric; the reported accuracy is an average of three different runs with seeds (40, 41, 42). For prototype interpretability evaluation we focus on various metrics depending on the benchmark:

\textbf{FunnyBirds} The FunnyBirds dataset is made up of synthetically generated bird images, each created by combining five distinct, human-interpretable features: beak, wings, feet, eyes, and tail, referred to as ”parts.” It includes $50$ bird categories, with each class representing
a unique subset of $26$ predefined parts.
The benchmark evaluates the methods on five different
dimensions: 
Accuracy (Acc.), Background Invariance (B.I.), Completness (Com.), Correctness (Cor.), Contrastivity (Con.).
For these experiments we do not use the transformations to align with how original models are trained and leverage the pre-trained checkpoints.

\textbf{Purity.} To evaluate the semantic quality of learned prototypes, we adopt the Purity metric following PiP-Net protocol, which quantifies the consistency between a model's latent representations and human-recognizable object parts. It is calculated as the fraction of a prototype’s top-10 most similar image patches that share the same ground-truth part annotation (e.g., "wing" or "wheel"). By measuring this alignment in range from 1 to 0, the metric reveals the "semantic gap" in a model, allowing for a rigorous comparison of whether its reasoning is based on coherent physical concepts or abstract, non-intuitive features.

\textbf{Spatial Misalignment}
This benchmark uses adversarial perturbations to modify the input image strictly outside the prototypical activation region, assessing how changes to "irrelevant" context affect both the local explanation and the model's global prediction.
\begin{itemize}
\item \textbf{Prototypical part Location Change (PLC)} – Measures the geometric instability of an explanation by calculating the shift in the activation bounding box after background modification; higher values indicate that the explanation's location depends on external context.
\item \textbf{Prototypical part Activation Change (PAC)} – Quantifies the relative decrease in a prototype's similarity score when pixels outside the highlighted region are perturbed, revealing the degree to which a "local" part's detection relies on global features.
\item \textbf{Prototypical part Rank Change (PRC)} – Measures the change in importance of a prototype relative to other classes; it counts how many out-of-class prototypes become more active than the target part following background modification.
\item \textbf{Accuracy Change (AC)} – Measures the drop in classification performance between original and modified images, determining if spatial misalignment is severe enough to cause the model to change its final behavior.
\end{itemize}

\section{FunnyBirds Results}
\label{sec:fb}
\begin{figure}
    \centering
    \includegraphics[width=0.8\linewidth]{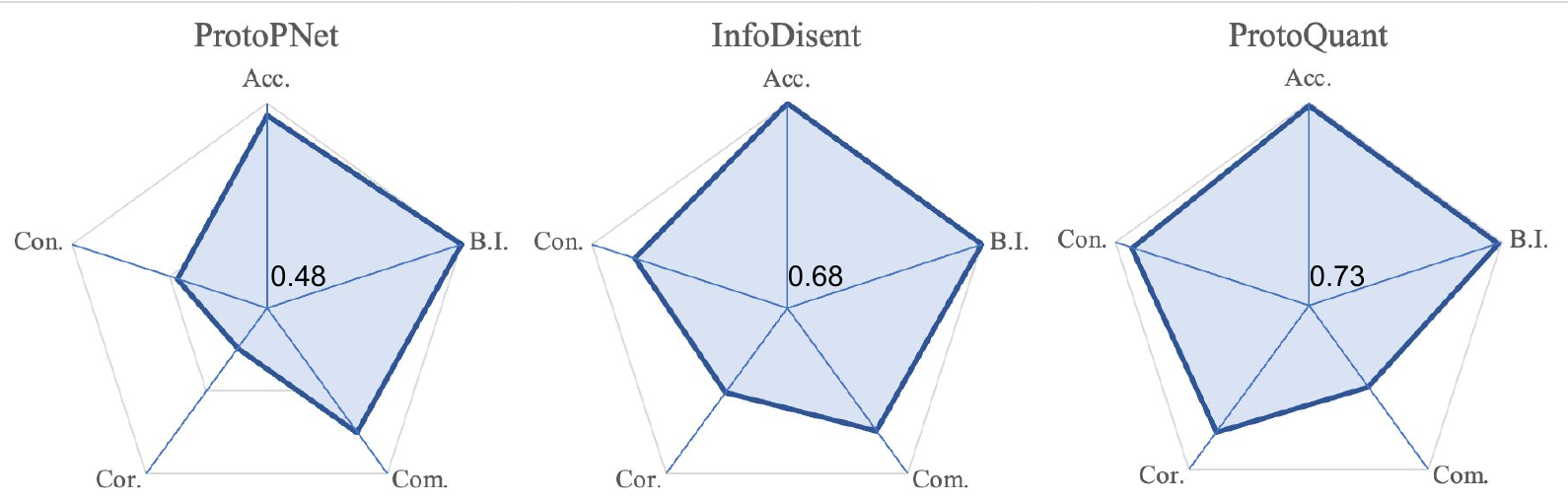}
    \caption{Here we present results on FunnyBirds benchmark when compared to other ante-hoc prototypical-parts based methods. One can observe that \our{} achieves the best results among tested architectures.}
    \label{fig:fb_results}
\end{figure}

We evaluate our method on the FunnyBirds dataset, which provides controlled, semantically meaningful image interventions, such as the removal or alteration of specific object parts. When compared to ProtoPNet and InfoDisent, \our{} achieves superior results (0.73 vs 0.48 and 0.68, respectively). This showcase that \our{} offers meaningful concept-level explanations. Detailed comparison can be found in the Figure~\ref{fig:fb_results}.

\section{Visualizations}
\label{sec:vis}

In this section we provide additional visualizations of explanations obtained from \our{}.

\begin{figure}[t]
    \centering
    \includegraphics[width=0.5\textwidth]{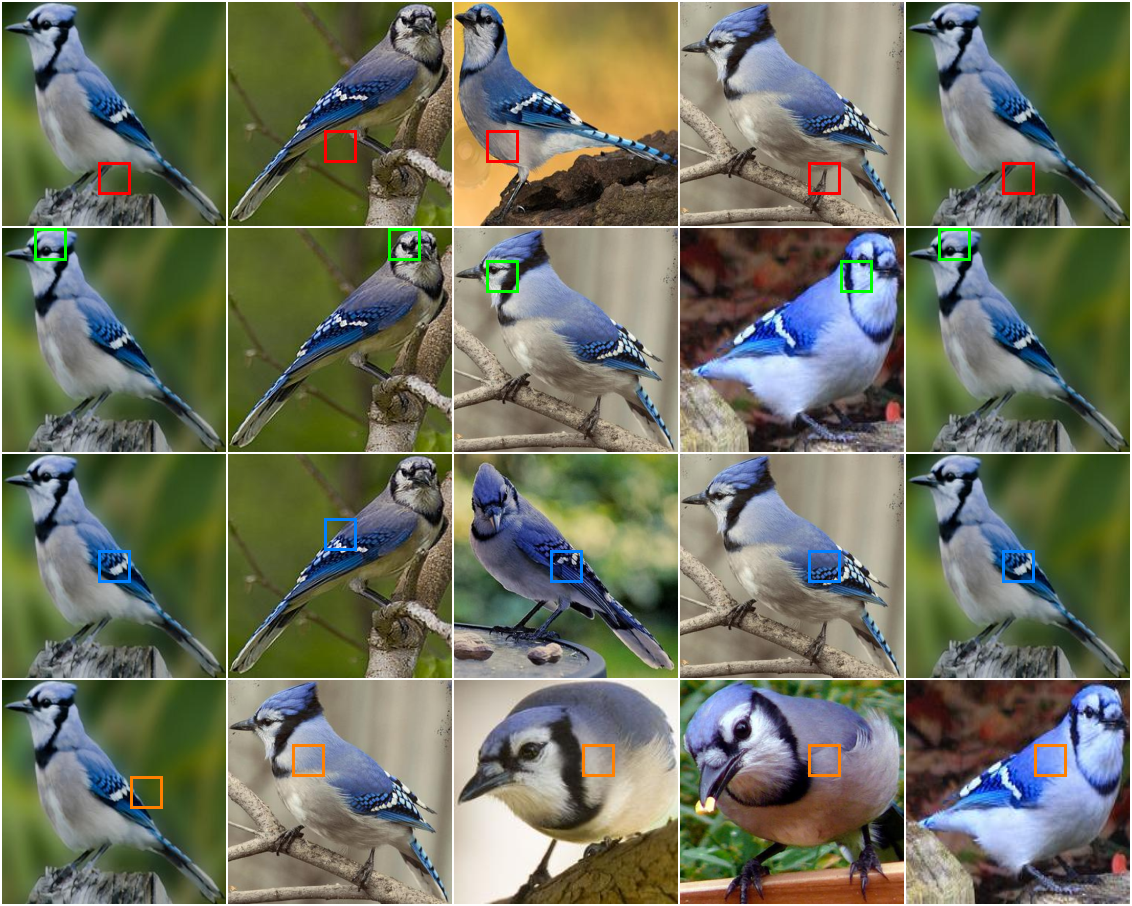}
    \caption{Visualization of the prototypes on CUB-200 dataset.}
    \label{fig:cub_vis_prototype2161}
    
\end{figure}

\begin{figure}[t]
\centering    \includegraphics[width=0.5\textwidth]{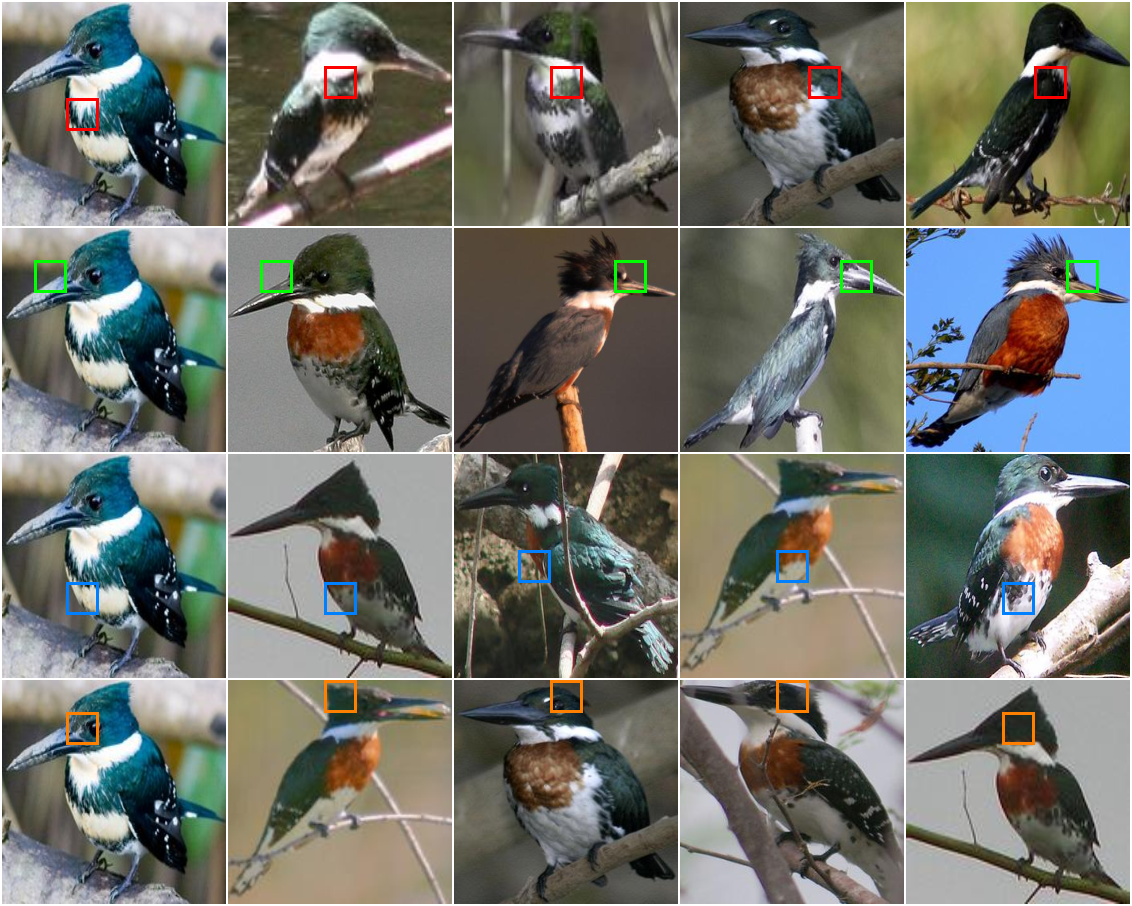}
    \caption{Visualization of the prototypes on CUB-200 dataset.}
    \label{fig:cub_vis_prototype2382}
    
\end{figure}

\begin{figure}[t]
\centering    \includegraphics[width=0.5\textwidth]{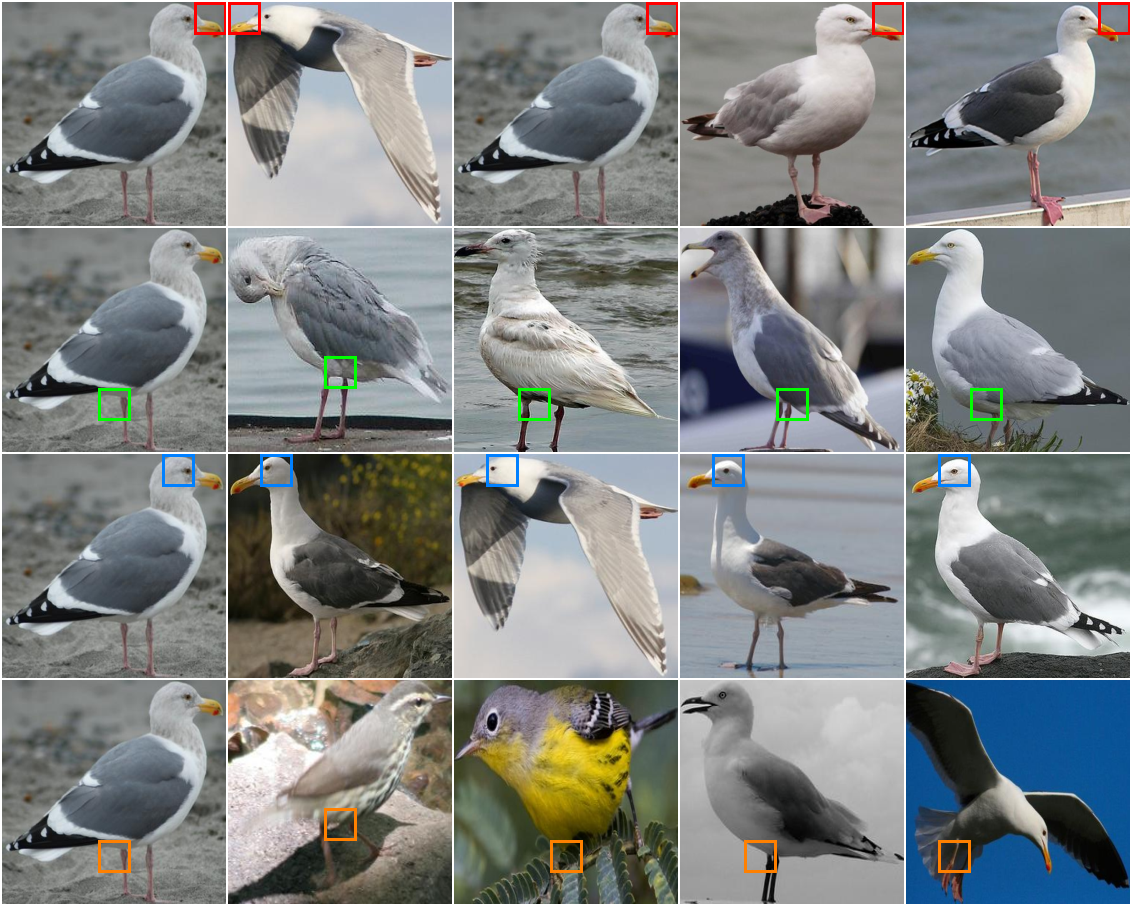}
    \caption{Visualization of the prototypes on CUB-200 dataset.}
    \label{fig:cub_vis_prototype2382}
\end{figure}

\begin{figure}[t]
\centering    \includegraphics[width=0.5\textwidth]{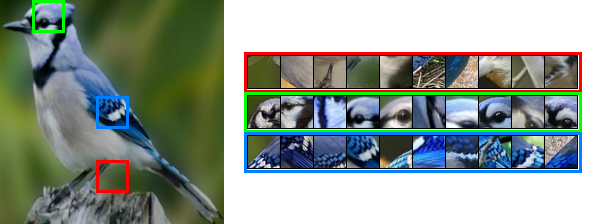}
    \caption{Visualization of the prototypes on CUB-200 dataset.}
    \label{fig:cub_vis_prototype2382}
\end{figure}

\begin{figure}[t]
\centering    \includegraphics[width=0.5\textwidth]{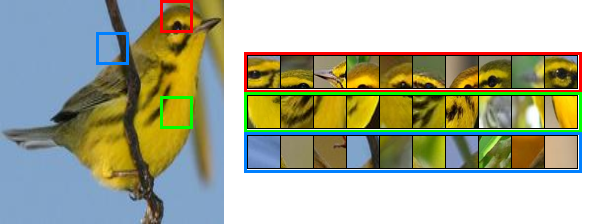}
    \caption{Visualization of the prototypes on CUB-200 dataset.}
    \label{fig:cub_vis_prototype2382}
\end{figure}

\begin{figure}[t]
\centering    \includegraphics[width=0.5\textwidth]{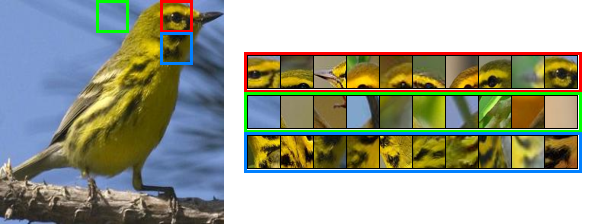}
    \caption{Visualization of the prototypes on CUB-200 dataset.}
    \label{fig:cub_vis_prototype2382}
\end{figure}

\begin{figure}[t]
\centering    \includegraphics[width=0.5\textwidth]{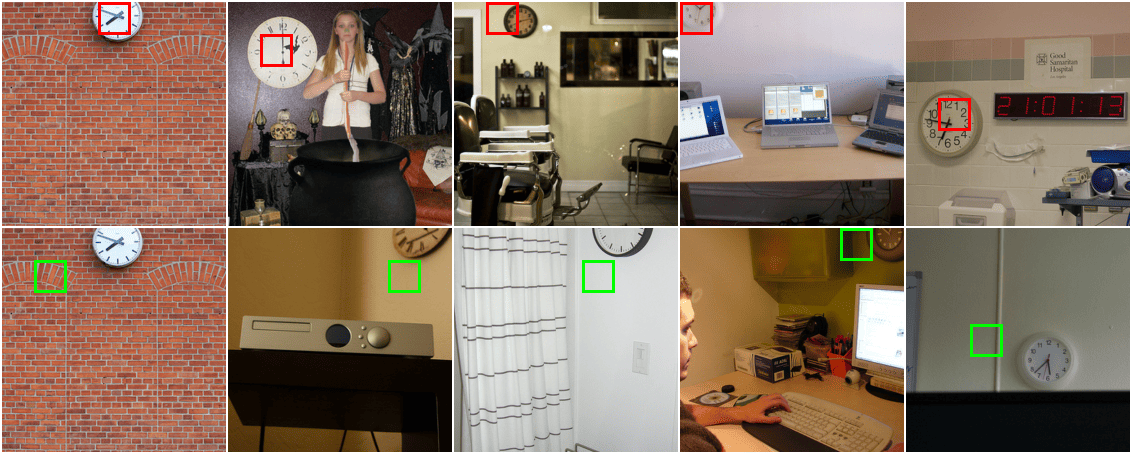}
    \caption{Visualization of the prototypes on ImageNet dataset, class clock.}
    \label{fig:cub_vis_prototype2382}
\end{figure}

\begin{figure}[t]
\centering    \includegraphics[width=0.5\textwidth]{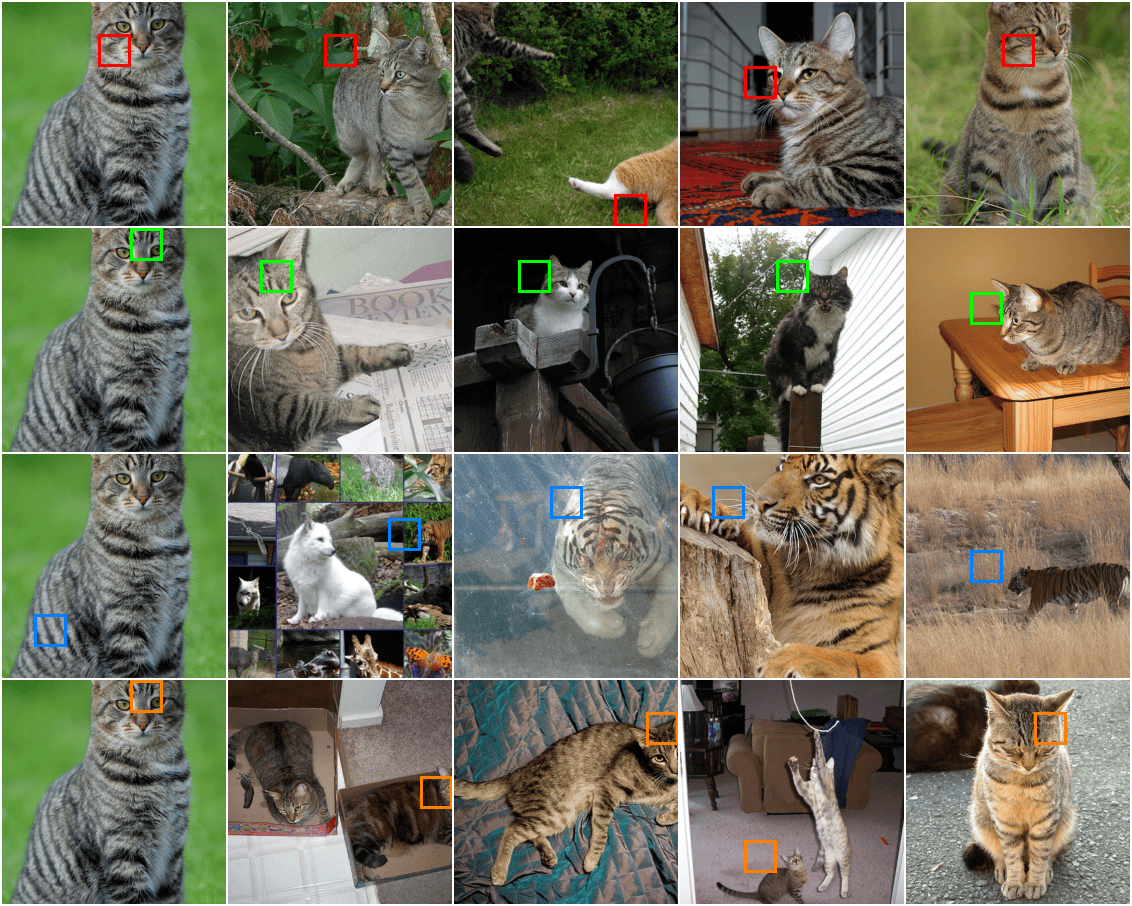}
    \caption{Visualization of the prototypes on ImageNet dataset, class cat.}
    \label{fig:cub_vis_prototype2382}
\end{figure}

\begin{figure}[t]
\centering    \includegraphics[width=0.5\textwidth]{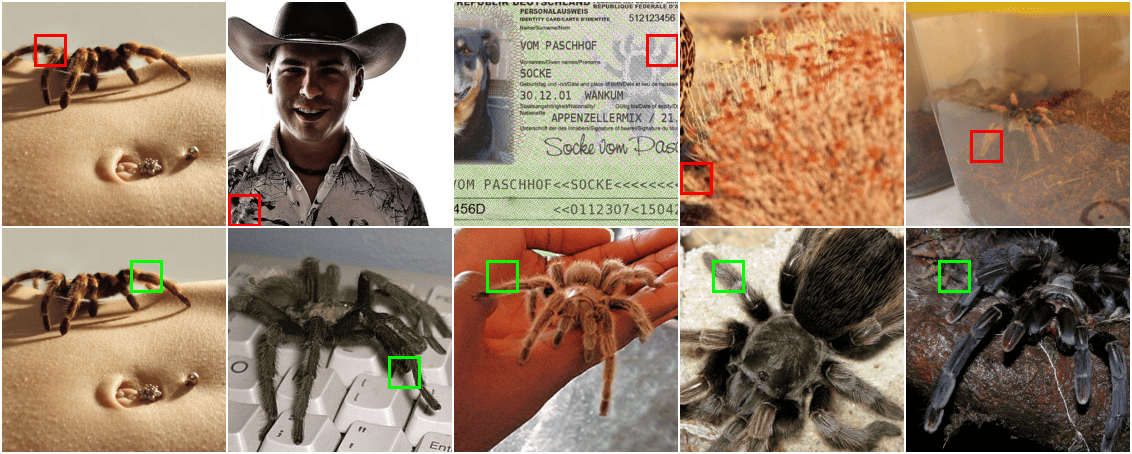}
    \caption{Visualization of the prototypes on ImageNet dataset, class spider.}
    \label{fig:cub_vis_prototype2382}
\end{figure}

\begin{figure}[t]
\centering    \includegraphics[width=0.5\textwidth]{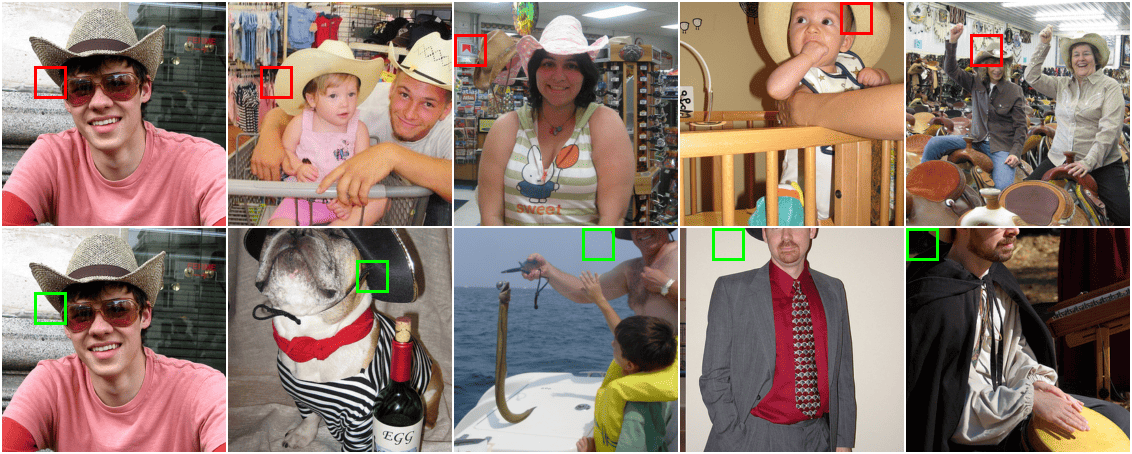}
    \caption{Visualization of the prototypes on ImageNet dataset, class hat.}
    \label{fig:cub_vis_prototype2382}
\end{figure}

\begin{figure}[t]
\centering    \includegraphics[width=0.5\textwidth]{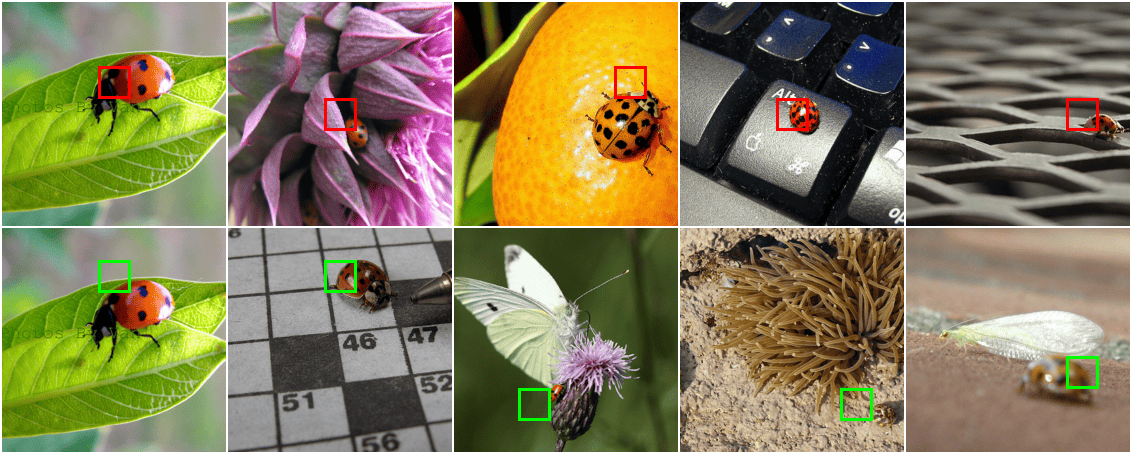}
    \caption{Visualization of the prototypes on ImageNet dataset, class ladybug.}
    \label{fig:cub_vis_prototype2382}
\end{figure}




\section{Detailed results}
\label{sec:dr}

Here we present detailed results in a tabular format (Table~\ref{tab:head_comparison} and ~\ref{tab:combined_sparsity} of plots presented in Figure~\ref{fig:codebook_size_ablation}) and Figure~\ref{fig:combined_sparsity}, respectively.

\begin{table*}[h]
\centering
\caption{Comparison of Top-1 Validation Accuracy between the Pretrained Head and the Interpretable Head (ProtoQuant) across various codebook sizes. Results represent the average of 3 seeds (40, 41, 42).}
\label{tab:head_comparison}
\begin{tabular}{@{}rrrr@{}}
\toprule
\textbf{Codebook Size} & \textbf{Pretrained Head} & \textbf{Interpretable Head} & \textbf{Difference} \\
& (\%) & (ProtoQuant, \%) & ($\Delta$) \\
\midrule
500    & 84.61 & 86.04 & +1.44 \\
1,000  & 86.62 & 86.92 & +0.31 \\
2,000  & 87.01 & 87.22 & +0.21 \\
4,000  & 87.41 & 87.67 & +0.26 \\
8,000  & 87.46 & 87.65 & +0.19 \\
16,000 & 87.63 & 87.67 & +0.04 \\
32,000 & 87.65 & 87.75 & +0.10 \\
64,000 & 87.89 & 87.72 & -0.17 \\
128,000& 87.79 & 87.44 & -0.35 \\
\bottomrule
\end{tabular}
\end{table*}

 \begin{table}[t]
 \caption{Sparsity vs. Representational Efficiency. Comparison of Top-1 accuracy on CUB-200 as we restrict the number of prototypes allowed per class ($K$). While the \textit{Frozen} Stage 2 model has higher peak capacity, the \textit{Finetuned} variant is significantly more efficient, maintaining full accuracy with only 10 prototypes per class showcasing that tuning of codebook after Stage 1 of training may be beneficial.}
 \label{tab:combined_sparsity}
 \vskip 0.15in
 \begin{center}
 \begin{small}
 \begin{sc}
 \begin{tabular}{lcccc}
 \toprule
 & \multicolumn{2}{c}{\textbf{Frozen Codes}} & \multicolumn{2}{c}{\textbf{Pruned \& Tuned Codes}} \\
 \cmidrule(r){2-3} \cmidrule(l){4-5}
 \textbf{$K$} & \textbf{Acc. (\%)} & \textbf{Codes} & \textbf{Acc. (\%)} & \textbf{Codes} \\
 \midrule
 Full & \textbf{87.76} & 3000 & - & - \\
 50   & 85.12 & 2724 & - & - \\
 30   & 82.78 & 2323 & - & - \\
 20   & 77.48 & 1954 & - & - \\
 10   & 67.21 & 1286 & \textbf{86.07} & \textbf{1498} \\
 8    & 62.72 & 1111 & 84.78 & 1260 \\
 5    & 56.14 & 786  & 78.24 & 786  \\
 1    & 31.26 & 185  & 42.03 & 182  \\
 \bottomrule
 \end{tabular}
 \end{sc}
 \end{small}
 \end{center}
 \vskip -0.1in
 \end{table}






\end{document}